\documentclass[conference]{IEEEtran}
\IEEEoverridecommandlockouts
\usepackage{cite}
\usepackage{amsmath,amssymb,amsfonts}
\usepackage{algorithmic}
\usepackage{graphicx}
\usepackage{textcomp}
\usepackage{xcolor}
\usepackage{subfigure}
\usepackage{algorithm}
\def\BibTeX{{\rm B\kern-.05em{\sc i\kern-.025em b}\kern-.08em
    T\kern-.1667em\lower.7ex\hbox{E}\kern-.125emX}}
\begin{document}

\title{Lazy Greedy Hypervolume Subset Selection from Large Candidate Solution Sets\\  
{\footnotesize}
\thanks{This work was supported by National Natural Science Foundation of China (Grant No. 61876075), Guangdong Provincial Key Laboratory (Grant No. 2020B121201001), the Program for Guangdong Introducing Innovative and Enterpreneurial Teams (Grant No. 2017ZT07X386), Shenzhen Science and Technology Program (Grant No. KQTD2016112514355531), the Program for University Key Laboratory of Guangdong Province (Grant No. 2017KSYS008).}
\thanks{Corresponding Author: Hisao Ishibuchi (hisao@sustech.edu.cn)}
}

\author{\IEEEauthorblockN{Weiyu Chen, Hisao Ishibuchi, Ke Shang}
\IEEEauthorblockA{Guangdong Provincial Key Laboratory of Brain-inspired Intelligent Computation, \\ Department of Computer Science and Engineering, Southern University of Science and Technology, \\Shenzhen 518055, China\\
11711904@mail.sustech.edu.cn, hisao@sustech.edu.cn, kshang@foxmail.com
}
}
 
\maketitle  

\begin{abstract}
Subset selection is a popular topic in recent years and a number of subset selection methods have been proposed. Among those methods, hypervolume subset selection is widely used. Greedy hypervolume subset selection algorithms can achieve good approximations to the optimal subset. However, when the candidate set is large (e.g., an unbounded external archive with a large number of solutions), the algorithm is very time-consuming.  In this paper, we propose a new lazy greedy algorithm exploiting the submodular property of the hypervolume indicator. The core idea is to avoid unnecessary hypervolume contribution calculation when finding the solution with the largest contribution. Experimental results show that the proposed algorithm is hundreds of times faster than the original greedy inclusion algorithm and several times faster than the fastest known greedy inclusion algorithm on many test problems. 
\end{abstract}


\section{Introduction}
Multi-objective optimization aims to optimize some potentially conflicting objectives simultaneously. In the past few decades, evolutionary multi-objective optimization (EMO) algorithms have shown promising performance in solving this kind of problem. Subset selection is a hot topic in the EMO area. It is involved in many phases of EMO algorithms. (i) In each generation, we need to select a pre-specified number of solutions from the current and offspring populations for the next generation. (ii) After the execution of EMO algorithms, the final population is usually presented to the decision-maker. However, if the decision-maker does not want to examine all solutions in the final population, we need to choose only a small number of representative solutions for the decision-makers. (iii) Since many good solutions are discarded during the execution of EMO algorithms \cite{1li2019empirical}, we can use an unbounded external archive (UEA) to store all non-dominated solutions examined during the execution of EMO algorithms. In this case, we need to select a subset of the UEA as the final result after their termination \cite{2ishibuchi2016compare,3tanabe2017benchmarking,4singh2018distance}.

Many subset selection methods have been proposed based on different selection criteria such as hypervolume-based subset selection \cite{5bader2011hype,6bringmann2014generic,7kuhn2016hypervolume,8guerreiro2017computing}, $\epsilon$-indicator-based subset selection \cite{9bringmann2014two} and distance-based subset selection \cite{3tanabe2017benchmarking}. Among these criteria, the hypervolume indicator has been widely used for subset selection \cite{5bader2011hype,6bringmann2014generic,7kuhn2016hypervolume,8guerreiro2017computing}. The hypervolume subset selection problem (HSSP) \cite{5bader2011hype} is to select a pre-specified number of solutions from a given candidate solution set to maximize the hypervolume of the selected solutions.

At present, the HSSP can only be efficiently solved in two dimensions. When the dimension is higher than two, the search for the exact optimal subset of the HSSP is NP-hard \cite{10rote2016selecting}. Some algorithms have been proposed to approximately solve the HSSP. They can be categorized into the following three classes: (i) hypervolume-based greedy inclusion, (ii) hypervolume-based greedy removal, and (iii) hypervolume-based genetic selection. These algorithms can achieve good approximations to the optimal subset.

However, when the candidate solution set is huge (e.g., tens of thousands of non-dominated solutions in a UEA) and/or the dimension is high (e.g., 10-objective problem), even greedy algorithms need long computation time. Some efficient algorithms (e.g., IHSO*\cite{11bradstreet2008fast} and IWFG\cite{12cox2016improving}) were proposed to quickly determine the solution with the least hypervolume contribution in each iteration of greedy removal algorithms. Guerreiro et al. \cite{8guerreiro2017computing} proposed an algorithm for efficiently updating the hypervolume contribution of each solution, which can reduce the runtime of greedy algorithms for the HSSP in up to four dimensions to polynomial time. Jiang et al. \cite{13jiang2014simple} also proposed an efficient mechanism for hypervolume contribution updating in any dimension to decrease the total runtime of a hypervolume-based EMO algorithm. 

In this paper, we propose a new greedy inclusion algorithm, which is applicable to large candidate solution sets with many objectives. This algorithm exploits the submodularity \cite{14nemhauser1978analysis} of the hypervolume indicator to reduce the unnecessary calculation of hypervolume contributions. Experimental results show that the proposed idea greatly improves the efficiency of greedy subset selection from large candidate solution sets of many-objective problems.

The rest of the paper is organized as follows. Section II describes the hypervolume indicator, hypervolume contribution and some related state-of-the-art algorithms. In section III, we describe our proposed algorithm in detail. Then in section IV, we show our experimental results where the proposed algorithm is compared with some state-of-the-art algorithms. Finally, we draw some conclusions in section V.

\section{Background}
\subsection{Hypervolume indicator and hypervolume contribution}
The hypervolume indicator \cite{15knowles2003bounded,16zitzler1998multiobjective} is a widely used metric to evaluate the diversity and convergence of a solution set. It is defined as the size of the objective space which is covered by a set of non-dominated solutions and bounded by a reference set \(R\). Formally, the hypervolume of a solution set \(S\) is defined as follows:
\begin{equation}
HV(S) := \int_{\mathbb{R}^d}^{} A_s(x)dx,
\end{equation}
where \(d\) is the number of dimension and $A_s$ is the attainment function of $S$ with respect to the reference set $R$ and can be written as
\begin{equation}
A_s(x) =\left\{
\begin{aligned}
1 &          &if\ \exists \  s \in S, r \in R : f(s) \leq x \leq r,\\
0 &          & otherwise.
\end{aligned}
\right.
\end{equation}

Calculating the hypervolume of a solution set is a \#P-hard problem \cite{17bringmann2010approximating}. A number of algorithms have been proposed to quickly calculate the exact hypervolume such as Hypervolume by Slicing Objectives (HSO) \cite{18durillo2010jmetal,19durillo2011jmetal}, Hypervolume by Overmars and Yap (HOY) \cite{20,21overmars1991new,22beume2009s}, and Walking Fish Group (WFG) \cite{23while2011fast}. Among those algorithms, WFG has been generally accepted as the fastest one. 
The hypervolume contribution is defined based on the hypervolume indicator. The hypervolume contribution of a point \(p\) to a set \(S\) is
\begin{equation}
HVC(p, S) = HV(S \cup \{p\}) - HV(S).
\end{equation}

Fig. 1 illustrates the hypervolume of a solution set and the hypervolume contribution of a solution to the solution set in two dimensions. The grey region is the hypervolume of the solution set \( S = \{a,b,c,d,e\} \) and the yellow region is the hypervolume contribution of a solution \(p\) to \(S\).

\begin{figure}[htbp]
\centering
\includegraphics[width= 0.44\textwidth]{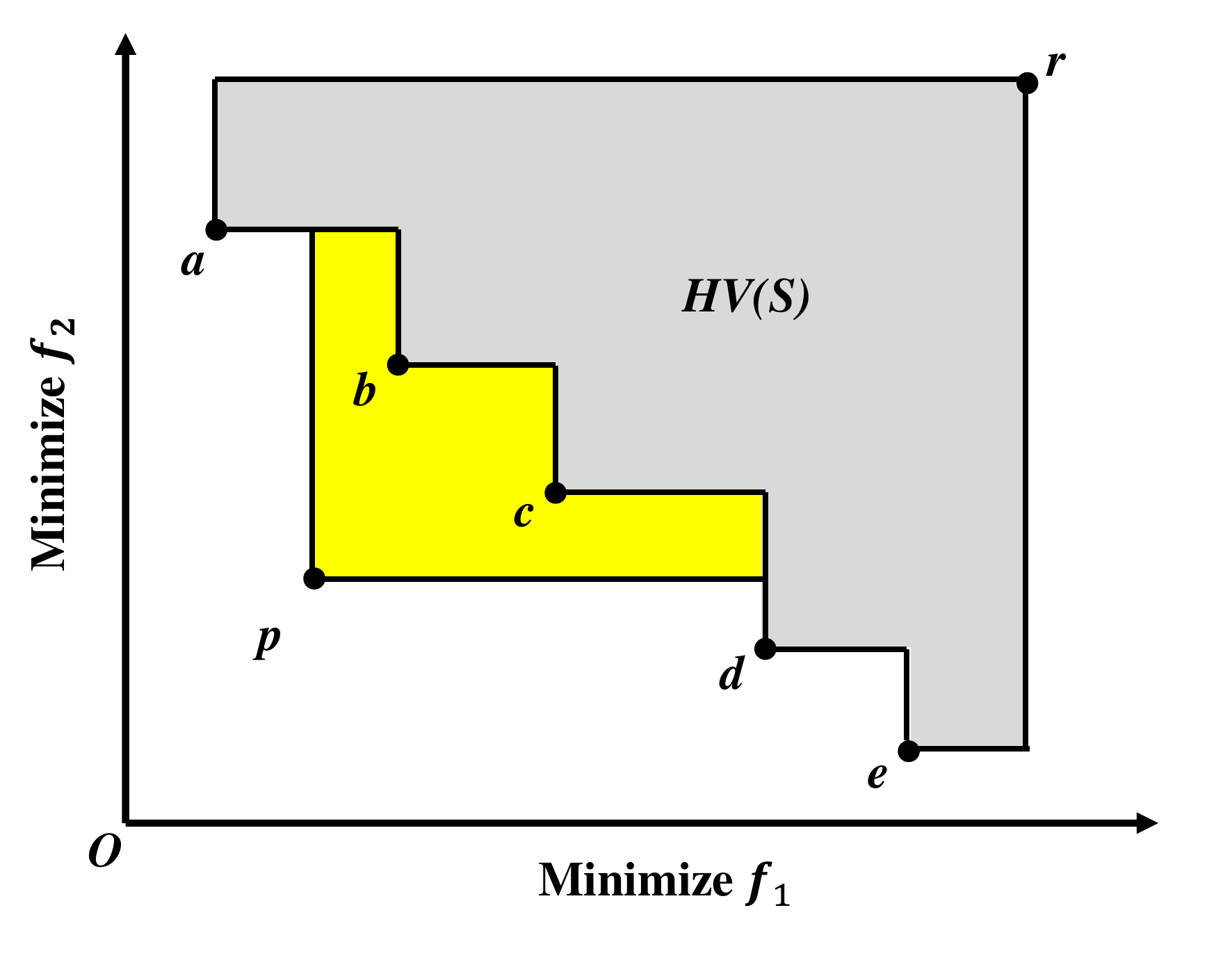}
\caption{The hypervolume of the solution set $S = \{a,b,c,d,e\}$ and the hypervolume contribution of \(p\) to the solution set \(S\) for a two-objective minimization problem.}
\end{figure}

Note that calculating the hypervolume contribution based on its definition in (3) requires hypervolume calculation twice, which is not very efficient. Bringmann and Friedrich \cite{24bringmann2009approximating} and Bradstreet et al. \cite{25bradstreet2009new} proposed a new calculation method to reduce the amount of calculation. The hypervolume contribution is calculated as
\begin{equation}
HVC(p, S) = HV(\{p\}) - HV(S'),
\end{equation}
where
\begin{equation}
S' = \{limit(s,p)|s \in S\},
\end{equation}
\begin{equation}
\begin{split}
&limit((s_1, ..., s_d), (p_1, ..., p_d)) \\
&= (worse(s_1, p_1),...,worse(s_d, p_d)).
\end{split}
\end{equation}

In this formulation $worse\left(s_i, p_i\right)$ takes the larger value. Compared to the straightforward calculation method in (3), this method is much more efficient. The hypervolume of one solution (i.e., \( HV\left(\{p\}\right)\)) can be easily calculated. We can also apply the previous mentioned HSO \cite{18durillo2010jmetal,19durillo2011jmetal}, HOY \cite{20,21overmars1991new,22beume2009s} and WFG \cite{23while2011fast} to calculate the hypervolume of a reduced solution set \(S’\) (i.e., \(HV\left(S'\right)\)).

Let us take Fig. 2 as an example. Suppose we want to calculate the hypervolume contribution of solution $p$ to a solution set $S = \{a,b,c,d,e\}$. First, for each solution in $S$, we replace each of its objective values with the corresponding value from solution $p$   if the value of $p$ is larger (i.e., we calculate $limit (a, p), ..., limit(e, p)$). This leads to $S’ = \{a’,b,c,d’,e’\}$. After the replacement, \(e’\) is dominated by \(d’\). Thus \(e’\) can be removed from \( S’\) since $e’$ has no contribution to the hypervolume of \(S’\). Then, we calculate the hypervolume of $S’$ (i.e., the area of the gray region in Fig. 2) and subtract it from the hypervolume of solution $p$. The remaining yellow part is the hypervolume contribution of solution $p$.
\begin{figure}[htbp]
\centering
\includegraphics[width=0.44 \textwidth]{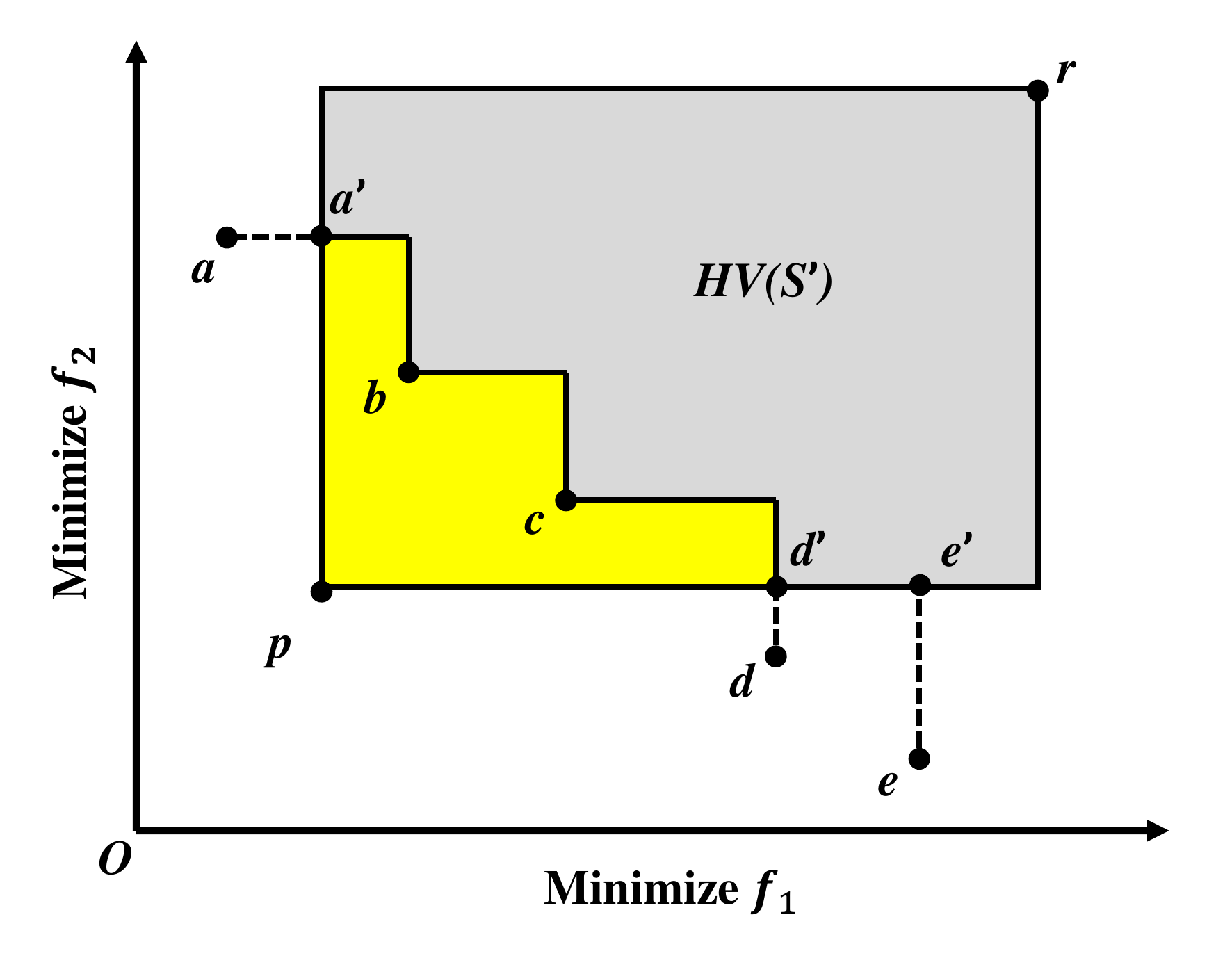}
\caption{Illustration of the efficient hypervolume contribution computation method.}
\end{figure}
\subsection{Hypervolume subset selection problem }
The hypervolume subset selection problem (HSSP) \cite{5bader2011hype}  is to select a pre-specified number (say \(k\)) of solutions from a given candidate solution set $S_{all}$ to maximize the hypervolume of the selected solutions (i.e., to select a subset \(S\) of size \(k\) from $S_{all}$ to maximize the hypervolume of \(S\)). Its formal definition is as follows.

Given an \(n\)-point set \(S_{all}\) and an integer \(k\) $\in$ \(\{0,1,…,|S_{all}|\}\), maximize \(HV(S)\) subject to \(S \subset S_{all}\) and \(|S| \leq  k\).

For two-objective problems, HSSP can be solved with time complexity of \(O(nk + nlogn) \) and \(O((n-k)k + nlogn)\) \cite{8guerreiro2017computing}. For multi-objective problems with three or more objectives, HSSP is an NP-hard problem \cite{10rote2016selecting}, it is impractical to try to find the exact optimal solution set when the size of the candidate set is large and/or the dimensionality of the objective space is high. In practice, some greedy heuristic algorithms and genetic algorithms are employed to obtain an approximated optimal solution set.

\subsection{Hypervolume-based greedy inclusion}
Hypervolume-based greedy inclusion selects solutions from $S_{all}$ one by one. In each iteration, the solution that has the largest hypervolume contribution to the selected solution set is selected until the required number of solutions are selected. The pseudocode of greedy inclusion is shown in Algorithm 1. The hypervolume-based greedy inclusion algorithm provides a \((1-1/e)\)-approximation ($e$ is the natural constant) to HSSP, which means the ratio of the hypervolume of the obtained solution set to the hypervolume of the optimal solution set is not less than \((1-1/e)\) \cite{14nemhauser1978analysis}. 

\begin{algorithm}
	\caption{Greedy Inclusion Hypervolume Subset Selection}
	\begin{algorithmic}[1]
    \REQUIRE  $S_{all}$ (A set of non-dominated solutions), $k$ (Solution subset size)
    \ENSURE $S$  (The selected subset from $S_{all}$)
     \IF{$|S_{all}| < k $}
           \STATE $S = S_{all}$
     \ELSE
  			\STATE  $S = \emptyset$
    		\WHILE{$|S| < k $}
    			\FOR {\textbf{each} $s_i$ in $S_{all} \setminus S$}
         				\STATE calculate the hypervolume contribution of $s_i$ to $S$
                 \ENDFOR
      			\STATE $p$ = solution in  $S_{all}\setminus S$  with the largest hypervolume contribution
      			\STATE $S = S \cup \{p\}$ 
             \ENDWHILE  
      \ENDIF
	\end{algorithmic}
\end{algorithm}

\subsection{Hypervolume-based greedy removal}
In contrast to greedy inclusion algorithms, hypervolume-based greedy removal algorithms discard one solution with the least hypervolume contribution to the current solution set in each iteration. To quickly identify the solution with the least hypervolume contribution, Incremental Hypervolume by Slicing Objectives (IHSO*) \cite{11bradstreet2008fast} and Incremental WFG (IWFG) \cite{12cox2016improving} were proposed. These methods can be used in the greedy removal algorithm. Some experimental results show that these methods can greatly accelerate greedy removal algorithms.

Unlike greedy inclusion, greedy removal has no approximation guarantee. It can obtain an arbitrary bad solution subset \cite{31bringmann2010efficient}. However, in practice, it usually leads to good approximations.

When the required set size \(k\) is close to the size of $S_{all}$ (i.e., when the number of solutions to be removed is small), greedy removal algorithms are faster than greedy inclusion algorithms. However, when \(k\) is relatively small in comparison with the size of $S_{all}$, greedy removal algorithms are not efficient since it needs to remove a large number of solutions.

\subsection{Hypervolume contribution update}
Hypervolume-based greedy inclusion/removal algorithms can be accelerated by updating hypervolume contributions instead of recalculating them in each iteration (i.e., by utilizing the calculation results in the previous iteration instead of calculating hypervolume contributions in each iteration independently). Guerreiro et al. \cite{9bringmann2014two} proposed an algorithm to update the hypervolume contributions efficiently in three and four dimensions. Using their algorithm, the time complexity of hypervolume-based greedy removal in three and four dimensions can be reduced to \(O (n(n - k) + n log n) \) and \(O(n^2 (n - k)) \) respectively.

In a hypervolume-based EMO algorithm called FV-MOEA proposed by Jiang et al.\cite{13jiang2014simple}, an efficient hypervolume contribution update method applicable to any dimension was proposed. The main idea of their method is that the hypervolume contribution of a solution is only associated with a small number of its neighboring solutions rather than all solutions in the solution set. Let us suppose that one solution \( s_j \) have just been removed from the solution set \( S\), the main process of the hypervolume contribution update method in \cite{13jiang2014simple} is shown in Algorithm 2. 

\begin{algorithm}
	\caption{Hypervolume Contribution Update}
	\begin{algorithmic}[1]
    \REQUIRE  $HVC$ (The hypervolume contribution of each solution in $S$), $s_j$ (The newly removed solution)
    \ENSURE $HVC$ (The updated hypervolume contribution of each solution in \(S\))
    \FOR {\textbf{each} $s_k \in S$}
         \STATE $w = worse(s_k, s_j)$
         \STATE $W = limit(S - \{s_k\}, w)$
         \STATE $HVC(s_k) = HVC(s_k) + HV(\{w\}) - HV(W)$
    \ENDFOR 
	\end{algorithmic}
\end{algorithm}
The \textit{worse} and \textit{limit} operations in Algorithm 2 are the same as those in Section II-A. Let us explain the basic idea of Algorithm 2 using Fig. 3. When we have a solution set \(S = \{a, b, c, d, e\}\) in Fig. 3, the hypervolume contribution of solution \( c\) is the blue area. When solution \(b\) is removed, the hypervolume contribution of \(c\) is updated as follows. The worse solution \(w\) in line 2 of Algorithm 2 has the maximum objective values of solutions \(b\) and \(c\). In line 3, firstly the limit operator changes solutions \(a\), \(d\) and \(e\) to \(a’\), \(d’\) and \(e’\). Next, the dominated solution \(e’\) is removed.  Then the solution set \(W = \{a’,d’\}\) is obtained. In line 4, the hypervolume contribution of \(c\) is updated by adding the term \(HV(\{w\})-HV(W)\) to its original value (i.e., the blue region in Fig. 3). The added term is the joint hypervolume contribution of solutions \(b\) and \( c\) (i.e., the yellow region in Fig. 3).  In this way, the hypervolume contribution of each solution is updated.

Since the \textit{limit} process reduces the number of non-dominated solutions, this updated method greatly improves the speed of hypervolume-based greedy removal algorithms. Algorithm 2 in \cite{13jiang2014simple} is the fastest known algorithm to update the hypervolume contribution in any dimension.

\begin{figure}[htbp]
\centering
\includegraphics[width=0.48 \textwidth]{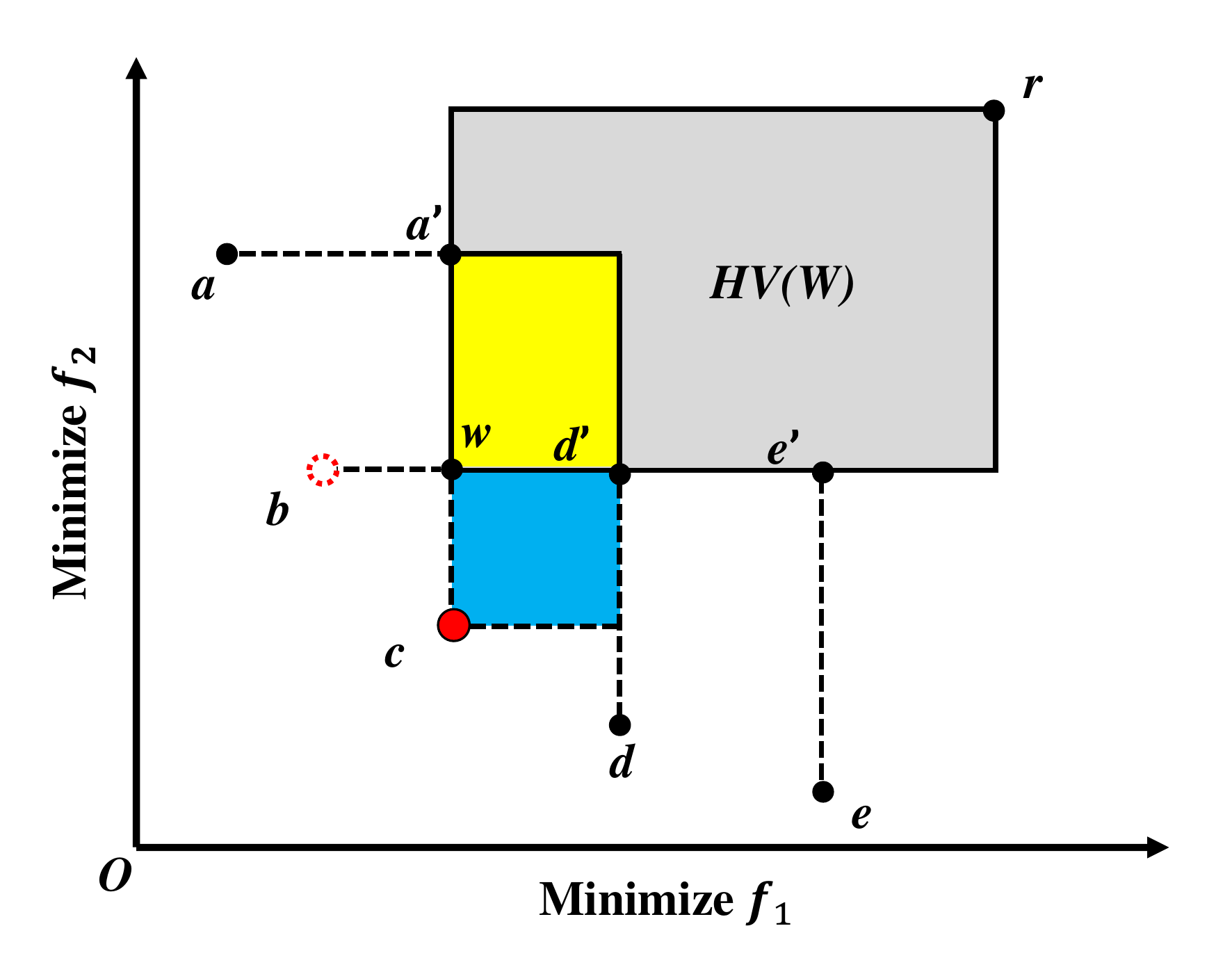}
\caption{Illustration of the hypervolume contribution update method in FV-MOEA. In this figure, it is assumed that point \(b\) has just been removed and the hypervolume contribution of point \(c\) is to be updated. }
\end{figure}
\section{Lazy Greedy Subset Selection Algorithm}
\subsection{Algorithm proposal}
In each iteration of hypervolume-based greedy inclusion algorithms, we only need to identify the solution with the largest hypervolume contribution. However, we usually calculate the hypervolume contributions of all solutions. Since it is time-consuming to calculate the hypervolume contribution of each solution, such an algorithm is not efficient. The main idea of the proposed algorithm is to exploit the submodular property of the hypervolume indicator \cite{27ulrich2012bounding}. The definition of a submodular function \cite{14nemhauser1978analysis} is as follows.

Given a finite nonempty set $N$, a real-valued function \(z(S)\) defined on the set of all subsets of \(N\) that satisfies 
$$z(S\cup\{k\})-z(S) \leq z(R \cup \{k\})-z\{R\},$$
$$ R\subset S\subset N, k\in N - S$$
 is called a submodular function.
 
The hypervolume indicator is a submodular function\cite{27ulrich2012bounding}. It means that the hypervolume contribution of a solution to the selected solution subset \(S\) never increases as the number of solutions in \(S\) increases in a greedy inclusion manner. Hence, instead of recomputing the hypervolume contribution of every candidate solution in each iteration, we can utilize the following lazy evaluation mechanism. We use a list \(C\) to store the candidate (i.e., unselected) solutions and their tentative HVC (hypervolume contribution) values. The tentative HVC value of each solution is initialized with its hypervolume (i.e., its hypervolume contribution when no solution is selected). The tentative HVC value of each solution is the upper bound of its true hypervolume contribution. For finding the solution with the largest hypervolume contribution from the list, we pick the most promising solution with the largest tentative HVC value, and recalculate its hypervolume contribution to the current solution subset \(S\). If the recalculated hypervolume contribution of this solution is still the largest in the list, we do not have to calculate the hypervolume contributions of the other solutions. This is because the hypervolume contribution of each solution never increases through the execution of greedy inclusion. In this case (i.e., if the recalculated hypervolume contribution of the most promising solution is still the largest in the list), we move this solution from the list to the selected solution subset \(S\). If the recalculated hypervolume contribution of this solution is not the largest in the list, its tentative HVC value is updated with the recalculated value. Then the most promising solution with the largest tentative HVC value in the list is examined (i.e., its hypervolume contribution is recalculated). This procedure is iterated until the recalculated hypervolume contribution is the largest in the list.

In many cases, the recalculation of the hypervolume contribution of each solution results in the same value as or a slightly smaller value than its tentative HVC value in the list since the inclusion of a single solution to the solution subset \(S\) changes the hypervolume contributions of only its neighbors in the objective space. Thus, the solution with the largest hypervolume contribution is often found without examining all solutions in the list. By applying this lazy evaluation mechanism, we can avoid a lot of unnecessary calculations in hypervolume-based greedy inclusion algorithms.

Since we always need to find the largest tentative HVC value in \(C\), the priority queue implemented by the maximum heap is used to accelerate the procedure. The details of the proposed a lazy greedy inclusion hypervolume-based subset selection (LGI-HSS) algorithm are shown in Algorithm 3.

The idea of the lazy evaluation was proposed by Minoux \cite{26minoux1978accelerated} to accelerate the greedy algorithm for maximizing submodular functions. Then, it was applied to some specific areas such as influence maximization problems \cite{30leskovec2007cost}. Minoux \cite{26minoux1978accelerated} proved that if the function is non-decreasing submodular and the greedy solution is unique, the solution produced by the lazy greedy algorithm and the original greedy algorithm is identical. Since it is proved that the hypervolume indicator is non-decreasing submodular \cite{27ulrich2012bounding}, the LGI-HSS algorithm will obtain the same subset as the original greedy inclusion algorithm if they use the same tie-break mechanism.

\begin{algorithm}
	\caption{Lazy Greedy Inclusion Hypervolume Subset Selection (LGI-HSS)}
	\begin{algorithmic}[1]
    \REQUIRE  $S_{all}$ (A set of non-dominated solutions), $k$ (Solution subset size)
    \ENSURE $S$  (The selected subset from $S_{all}$)
     \IF{$|S_{all}| < k $}
           \STATE $S = S_{all}$
     \ELSE
  			\STATE  $S = \emptyset$,  $C = \emptyset$
  			\FOR {\textbf{each} $s_i$ in $S_{all}$}
  			    \STATE insert $\left(s_i, HV(\{s_i\})\right)$ into $C$
  			\ENDFOR
    		\WHILE{$|S| < k $}
    		    \WHILE{$C \not= \emptyset$}
    		    \STATE $c_{max}$ = solution with the largest HVC in $C$
    		    \STATE update the HVC of $c_{max}$ to $S$
    		    \IF{$c_{max}$ has the largest HVC in $C$}
    		        \STATE $S = S \cup \{c_{max}\}$
    		        \STATE $C = C \setminus \{c_{max}\}$
    		        \STATE \textbf{break}
    		    \ENDIF
    		    \ENDWHILE
             \ENDWHILE  
      \ENDIF
	\end{algorithmic}
\end{algorithm}

\subsection{An illustrative example}
Let us explain the proposed algorithm using a simple example. Fig. 4 shows the changes of the hypervolume contribution in list \(C\). The values in the parentheses are the stored HVC value of each solution to the selected subset. For illustration purposes, the solutions in the list are sorted by the stored HVC values. However, in the actual implementation of the algorithm, the sorting is not necessarily needed (especially when the number of candidate solutions is very large). This is because our algorithm only needs to find the most promising candidate solution with the largest HVC value in the list. 

Fig. 4 (i) shows the initial list $C$ including five solutions \(a\), \(b\), \(c\), \(d\) and \(e\). The current solution subset is empty. In Fig. 4 (i), solution \(a\) has the largest HVC value. Since the initial HVC value of each solution is the true hypervolume contribution to the current empty solution subset \(S\), no recalculation is needed. Solution \(a\) is moved from the list to the solution subset. 

In Fig. 4 (ii), solution \(b\) has the largest HVC value in the list after solution \(a\) is moved. Thus, the hypervolume contribution of \(b\) is to be recalculated. We assume that the recalculated HVC value is 4 as shown in Fig. 4 (iii).

Fig. 4 (iii) shows the list after the recalculation. Since the updated HVC value of \(b\) is not the largest, we need to choose solution \(e\) which has the largest HVC value in the list and recalculate its hypervolume contribution. We assume that the recalculated HVC value is 6 as shown in Fig. 4 (iv).

Fig. 4 (iv) shows the list after the recalculation. Since the recalculated HVC value of solution $e$ is still the largest in the list, solution \(e\) is moved from the list to the solution subset $S$. 
Fig. 4 (v) shows the list after the removal of \(e\). Solution \(c\) with the largest HVC value is examined.

In this example, when we select the second solution from the remaining four candidates (\(b\), \(c\), \(d\) and \(e\)), we evaluate the hypervolume contributions of only the two solutions (\(b\) and \(e\)). In the standard greedy inclusion algorithm, all four candidates are examined. In this manner, the proposed algorithm decreases the computation time of the standard greedy inclusion algorithm.

\begin{figure}[htbp]
\centering
\includegraphics[width=0.36\textwidth]{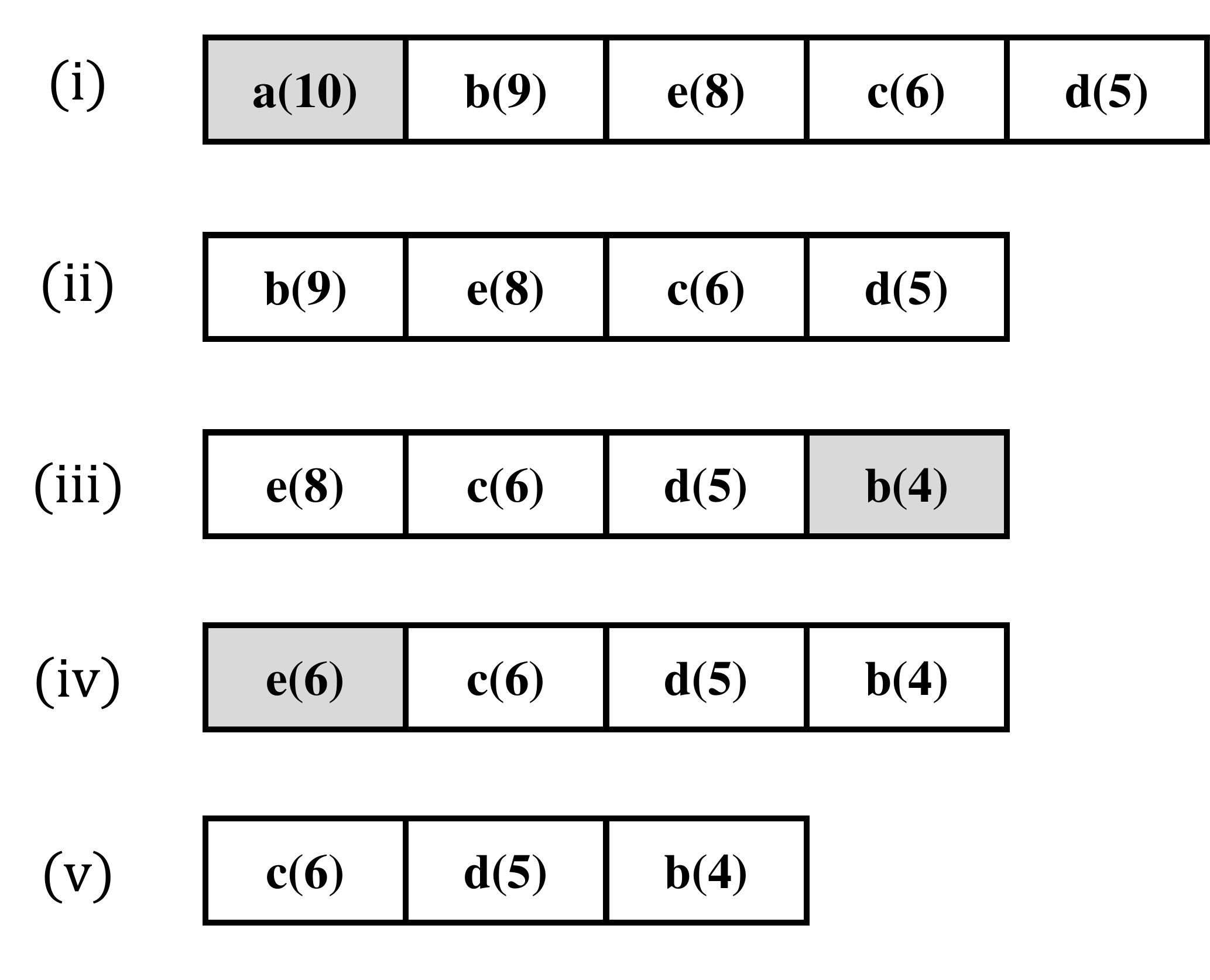}
\caption{Illustration of the proposed algorithm. The values in the parentheses are the stored tentative HVC values.}
\end{figure}
\section{Experiments}
\subsection{Algorithms for comparison}
The proposed LGI-HSS algorithm is compared with the following two algorithms:
 \begin{enumerate}
  \item Standard greedy inclusion hypervolume subset selection (GI-HSS): This is the greedy inclusion algorithm described in Section II-C. When calculating the hypervolume contribution, the effective method(i.e., formula (4)-(6)) described in Section II-A is employed.
  \item Greedy inclusion hypervolume subset selection with hypervolume contribution updating (UGI-HSS): The hypervolume contribution updating method proposed in FV-MOEA\cite{13jiang2014simple} (Algorithm 2) is used. Since Algorithm 2 is for greedy removal, it is changed for greedy inclusion here. It is the fastest known greedy inclusion algorithm applicable to any dimension.   
  \end{enumerate}
  
Since our main focus is the selection of a solution subset from an unbounded external archive (i.e., since the number of solutions to be selected is much smaller than the number of candidate solutions: $k << n$ in HSSP), greedy removal is not efficient. Hence, some algorithms only suitable for greedy removal (e.g., greedy removal using IHSO* \cite{11bradstreet2008fast} or IWFG \cite{12cox2016improving} to identify the least contribution solution) are not compared in this paper.

\subsection{Test Problems and Candidate Solutions}
To examine the performance of three subset selection algorithms, we choose three representative test problems with different Pareto front (PF) shapes:
\begin{enumerate}
\item Spherical front: Solutions on the true PF of the DTLZ2 test problem \cite{28deb2002scalable}. 
\item Discontinuous front: Solutions on the true PF of the DTLZ7 test problem \cite{28deb2002scalable}.
\item Inverted spherical front: Solutions on the true PF of the Inverted DTLZ2 (I-DTLZ2) problem \cite{29deb2013evolutionary}.
\end{enumerate}

For each test problem, we use three problem instances with 5, 8 and 10 objectives (i.e., solution subset selection is performed in five-, eight- and ten-dimensional objective spaces). Four different settings of the candidate solution set size are examined: 5000, 10000, 15000 and 20000. We first uniformly generate 100,000 solutions on the PF. In each run of a solution subset selection algorithm, a required number of candidate solutions (i.e., 5000, 10000, 15000 or 20000 solutions) are randomly selected from the generated 100,000 solutions for each problem instance. Computational experiments are performed five times for each setting of the candidate solution set size for each problem instance. The number of solutions to be selected is specified as 100. Thus our problem is to select 100 solutions from 5000, 10000, 15000 or 20000 candidate solutions to maximize the hypervolume of the selected solution.  
\subsection{Experimental settings}
In each subset selection algorithm, the reference point for hypervolume (contribution) calculation is set to $(1.1, 1.1, ..., 1.1)$ for all test problems independent of the number of objectives. We use the WFG algorithm \cite{23while2011fast} for hypervolume calculation in each solution subset selection algorithm. The code of the WFG algorithm is available from http://www.wfg.csse.uwa.edu.au/hypervolume/\#code. 

All subset selection algorithms are coded by MatlabR2018a. The computation time of each run is measured on an Intel Core i5-7200U CPU with 4GB of RAM, running in Windows 10. 
\subsection{Experimental results}
The results of the average computation time of each algorithm on the DTLZ2, DTLZ7 and I-DTLZ2 test problems are summarized in Figs. 5-7, respectively. Compared with the standard GI-HSS algorithm, we can see that our LGI-HSS algorithm can reduce the computation time by 91\% to 99\%. By the increase in the number of objectives (i.e., by the increase in the dimensionality of the objective space), the advantage of LGI-HSS over the other algorithms becomes larger. Among the three test problems in Figs. 5-7, all the three algorithms are fast on the I-DTLZ2 problem and slow on the DTLZ2 problem. 

Even when we compare our LGI-HSS algorithm with the fastest known greedy inclusion algorithm UDI-HSS , LGI-HSS is much faster. On DTLZ2 in Fig. 5, LGI-HSS spent 74\% to 96\% less computation time than UGI-HSS. On DTLZ7 in Fig. 6, LGI-HSS spent 47\% to 76\% less computation time than UGI-HSS. On the five-objective I-DTLZ2 problem instance in Fig. 7 (a), there is no large difference in the average computation time between the two algorithms (the average computation time of LGI-HSS is less than that of UGI-HSS by 34\%-58\%). However, by increasing the number of objectives in Fig. 7, the difference in the average computation time between the two algorithms becomes larger for I-DTLZ2. 

From Figs.  5-7, we can also observe that the average computation time of each algorithm did not severely increase when the number of objectives increases (i.e., when  the dimensionality of the objective space increases) for DTLZ7 in Fig. 6 and I-DTLZ2 in Fig. 7. In some cases, the average computation time of LGI-HSS decreased when the number of objectives increases (e.g., on I-DTLZ2 by LGI-HSS in Fig. 7). This issue needs to be further addressed in our future study.

\begin{figure}[htb]
\centering
  \subfigure[Five-objective DTLZ2 (5D spherical front).]{\includegraphics[width=0.478\textwidth]{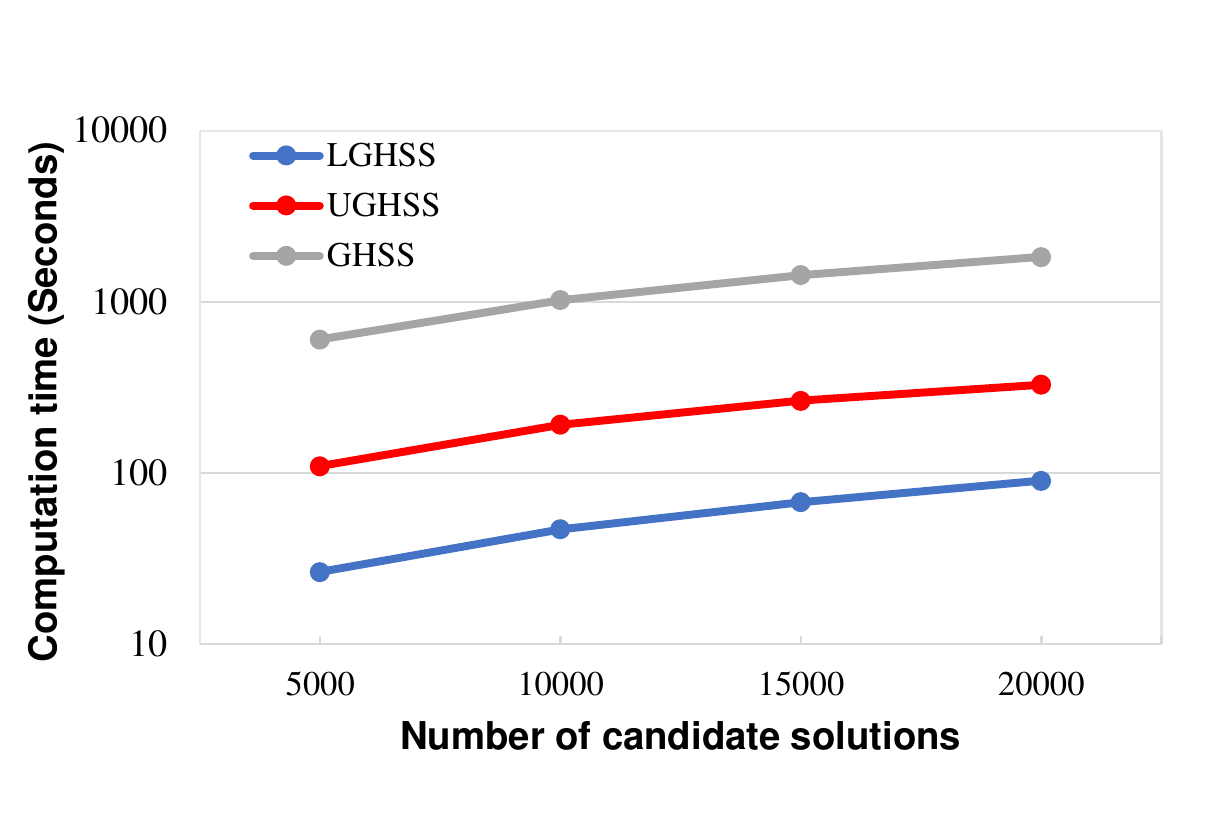}}
  \subfigure[Eight-objective DTLZ2 (8D spherical front).]{\includegraphics[width=0.478\textwidth]{8.pdf}}
  \subfigure[Ten-objective DTLZ2 (10D spherical front).]{\includegraphics[width=0.478\textwidth]{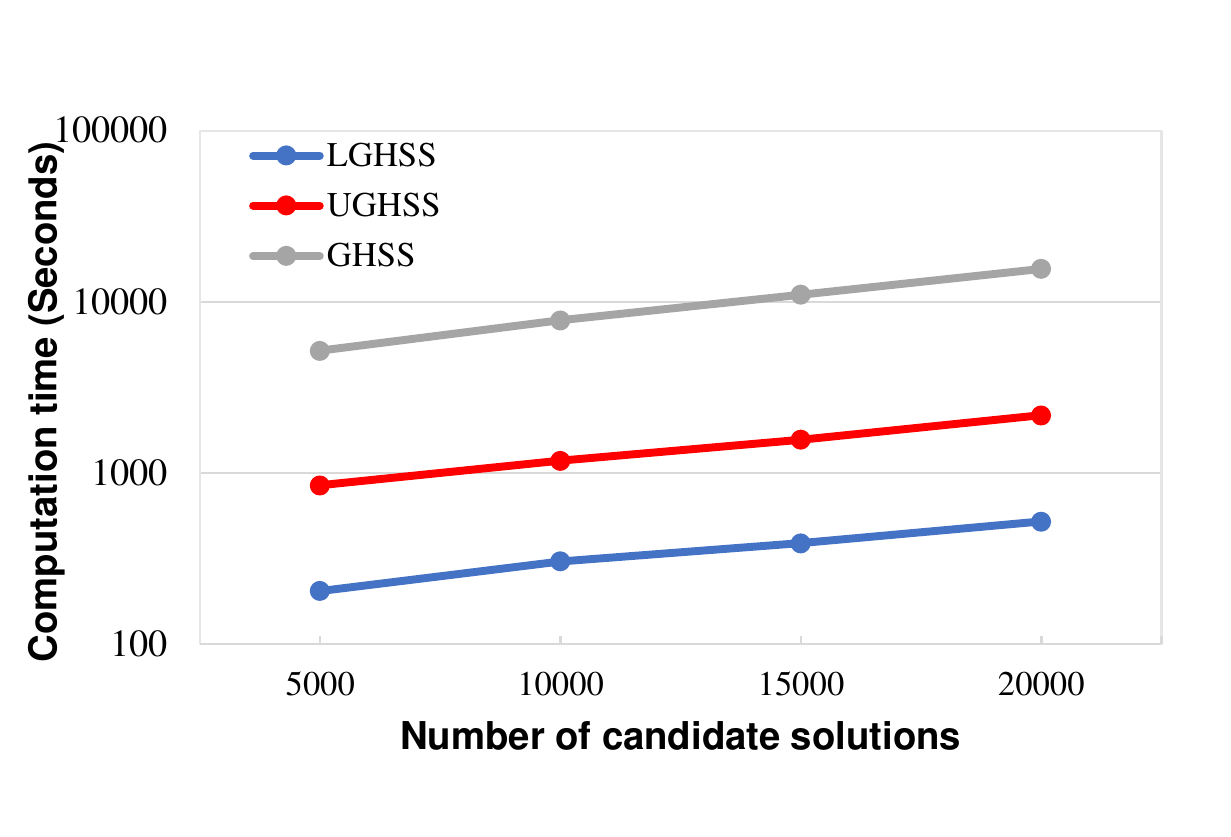}}
  \caption{ Average computation time on DTLZ2 with the spherical PF. The time axis is log scaled. }
\end{figure}

\begin{figure}[htb]
\centering
  \subfigure[Five-objective DTLZ7 (5D discontinuous front).]{\includegraphics[width=0.478\textwidth]{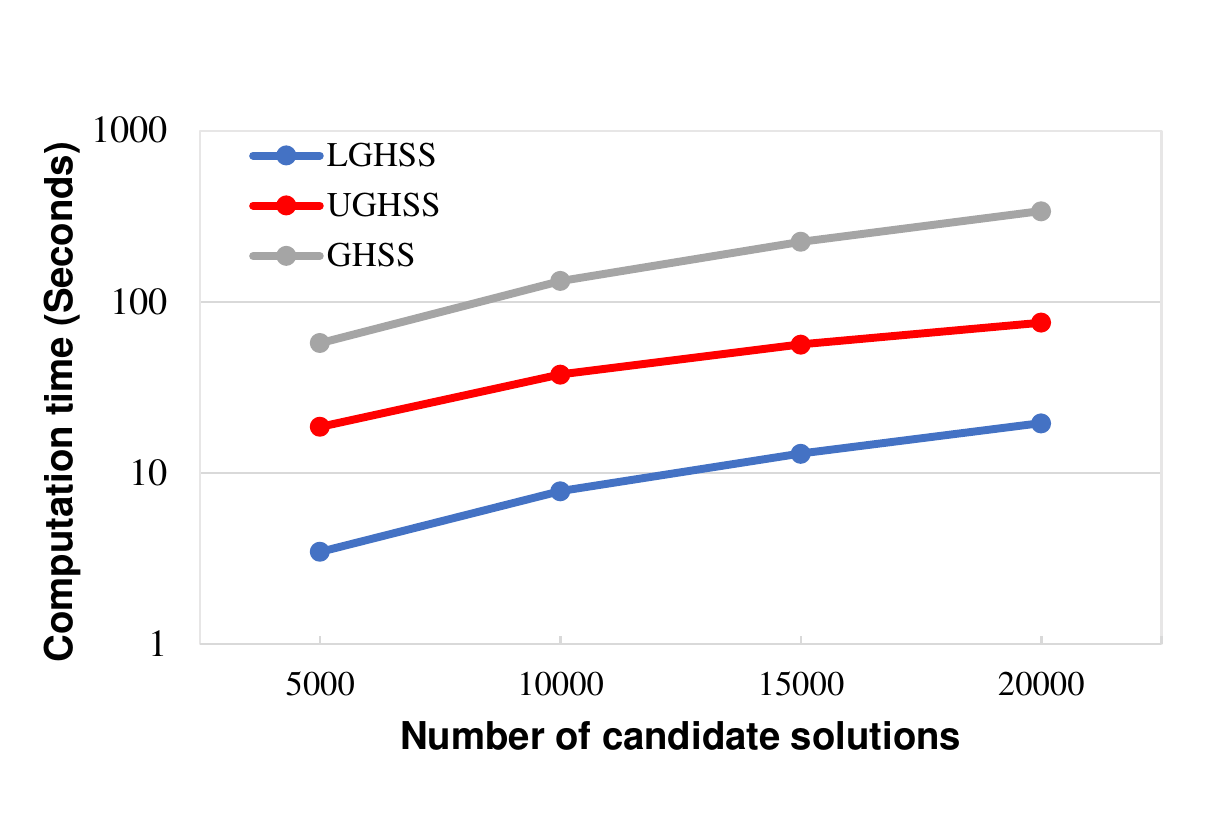}}
  \subfigure[Eight-objective DTLZ7 (8D discontinuous front).]{\includegraphics[width=0.478\textwidth]{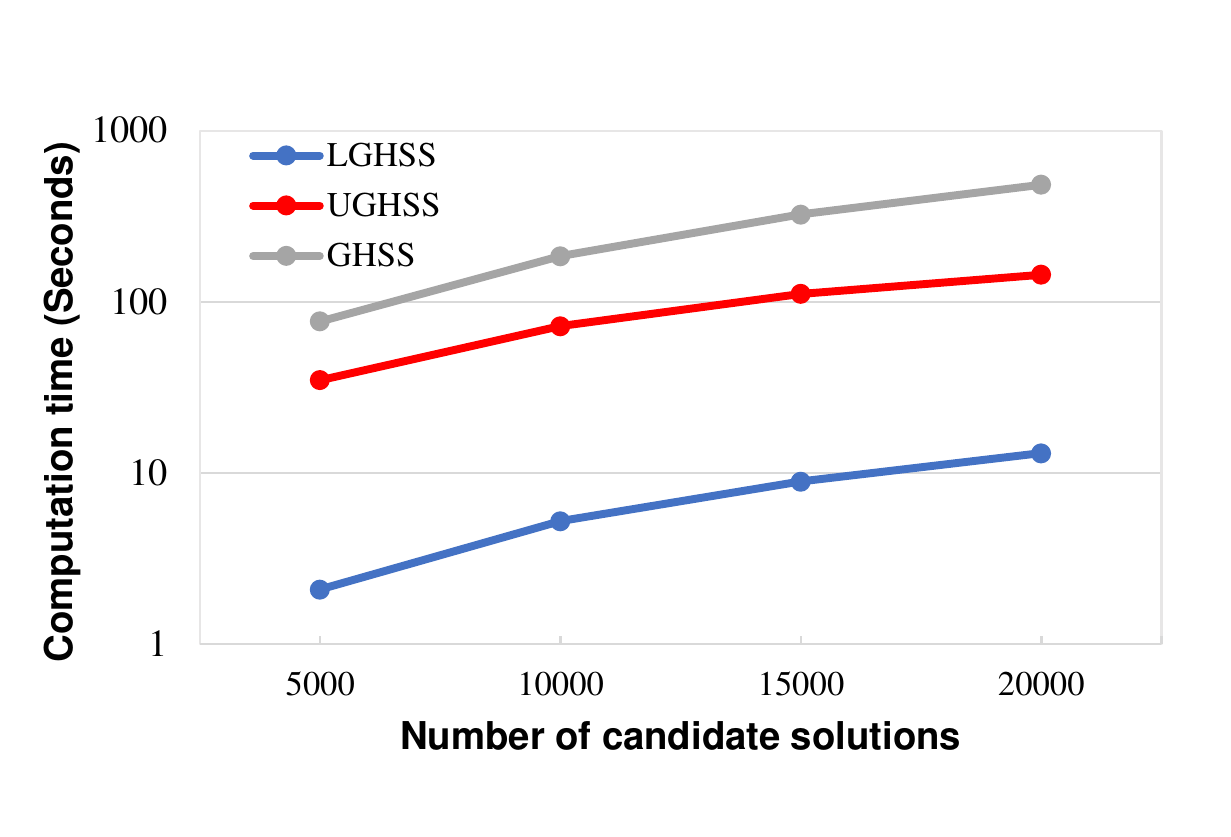}}
  \subfigure[Ten-objective DTLZ7 (10D discontinuous front).]{\includegraphics[width=0.478\textwidth]{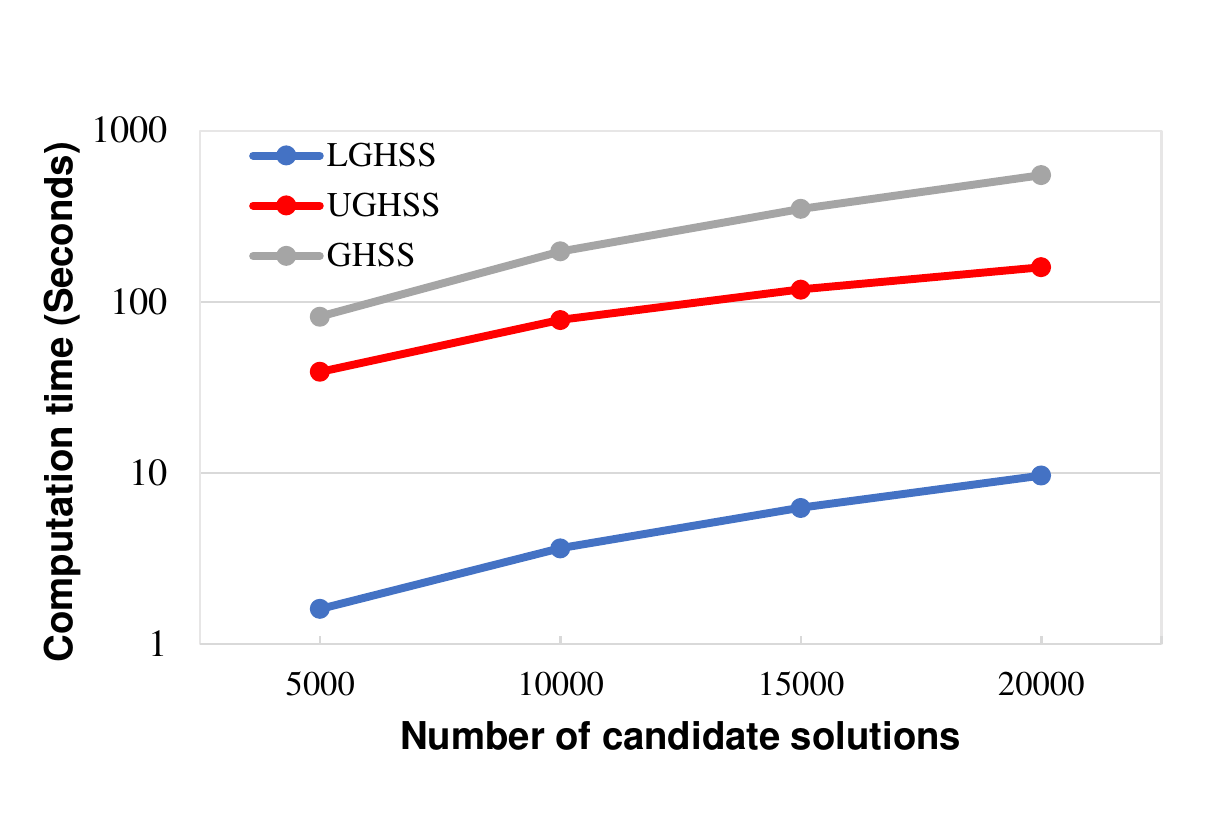}}
 
  \caption{ Average computation time on DTLZ7 with the discontinuous PF. The time axis is log scaled. }
\end{figure}

\begin{figure}[ht]
  \centering
  \subfigure[Five-objective I-DTLZ2 (5D inverted spherical front).]
  {\includegraphics[width=0.479\textwidth]{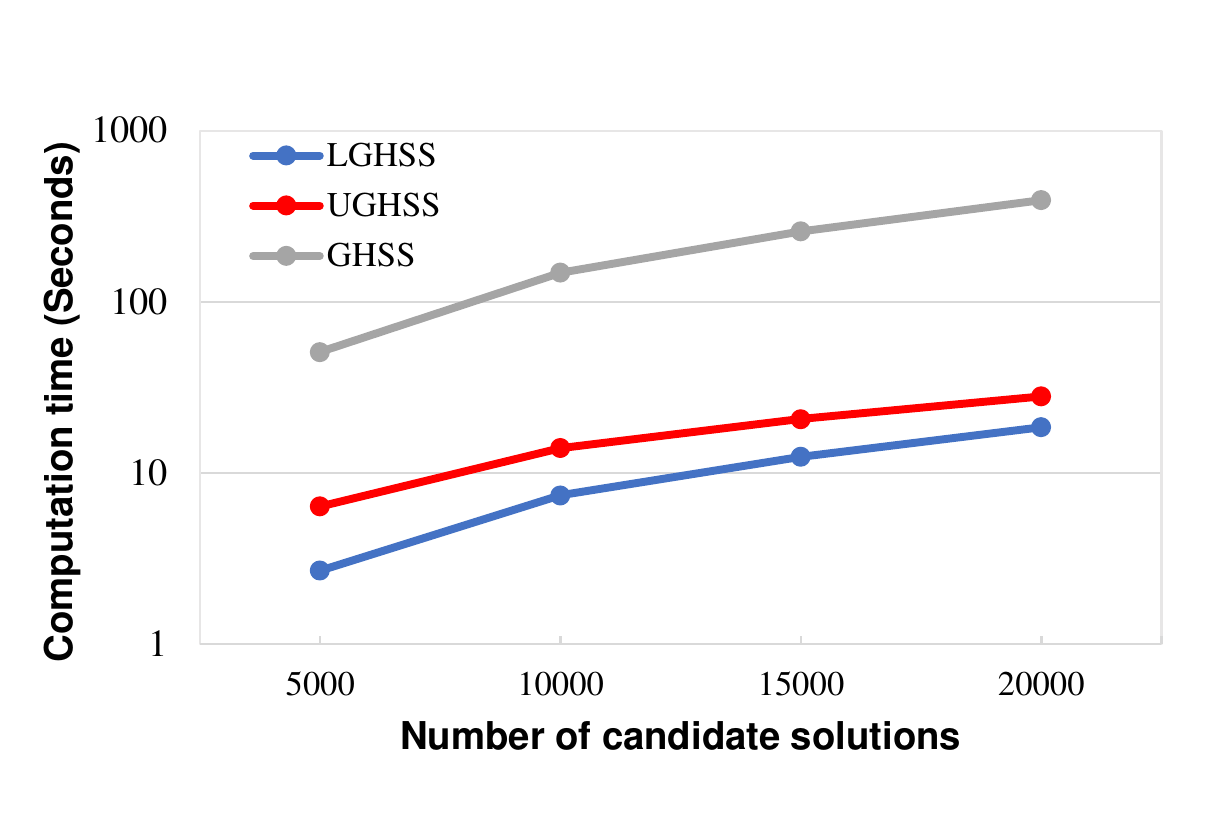}}
  \subfigure[Eight-objective I-DTLZ2 (8D inverted spherical  front).]
    {\includegraphics[width=0.479\textwidth]{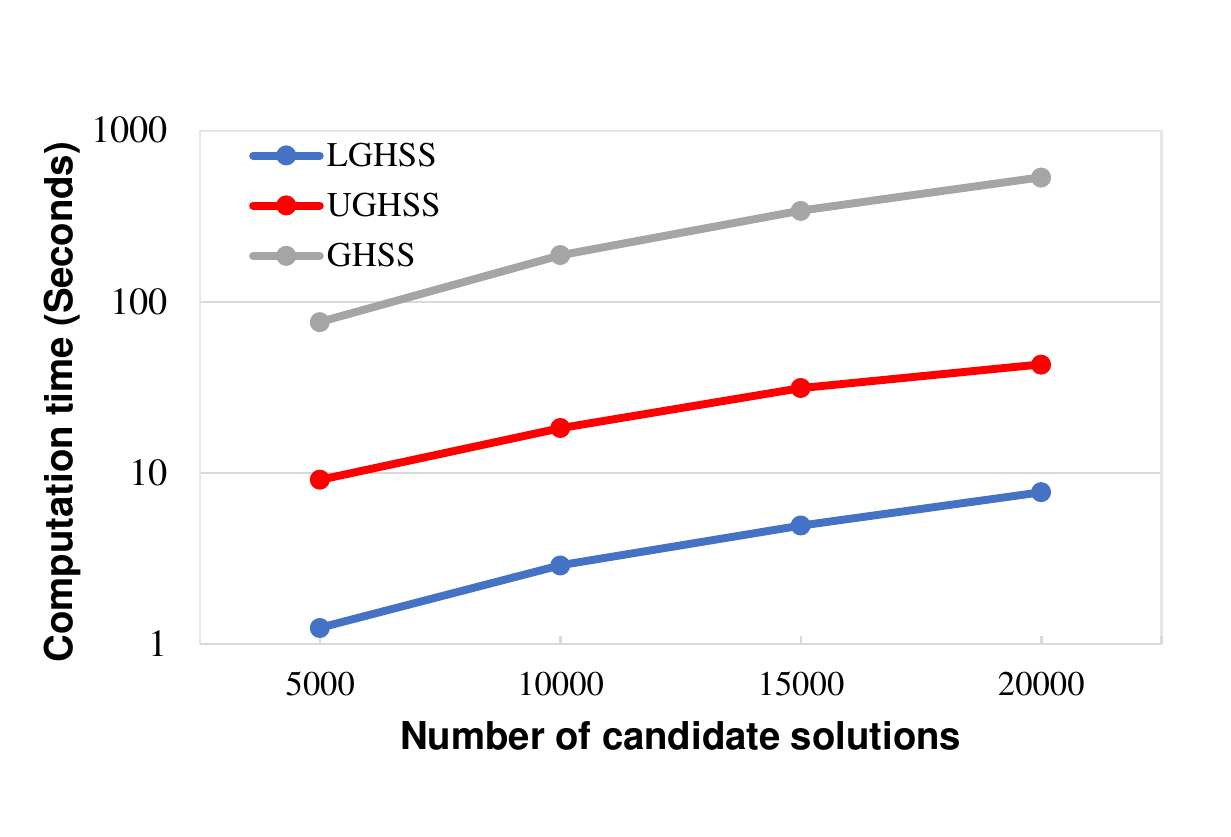}}
  \subfigure[Ten-objective I-DTLZ2 (10D inverted spherical  front).]
    {\includegraphics[width=0.479\textwidth]{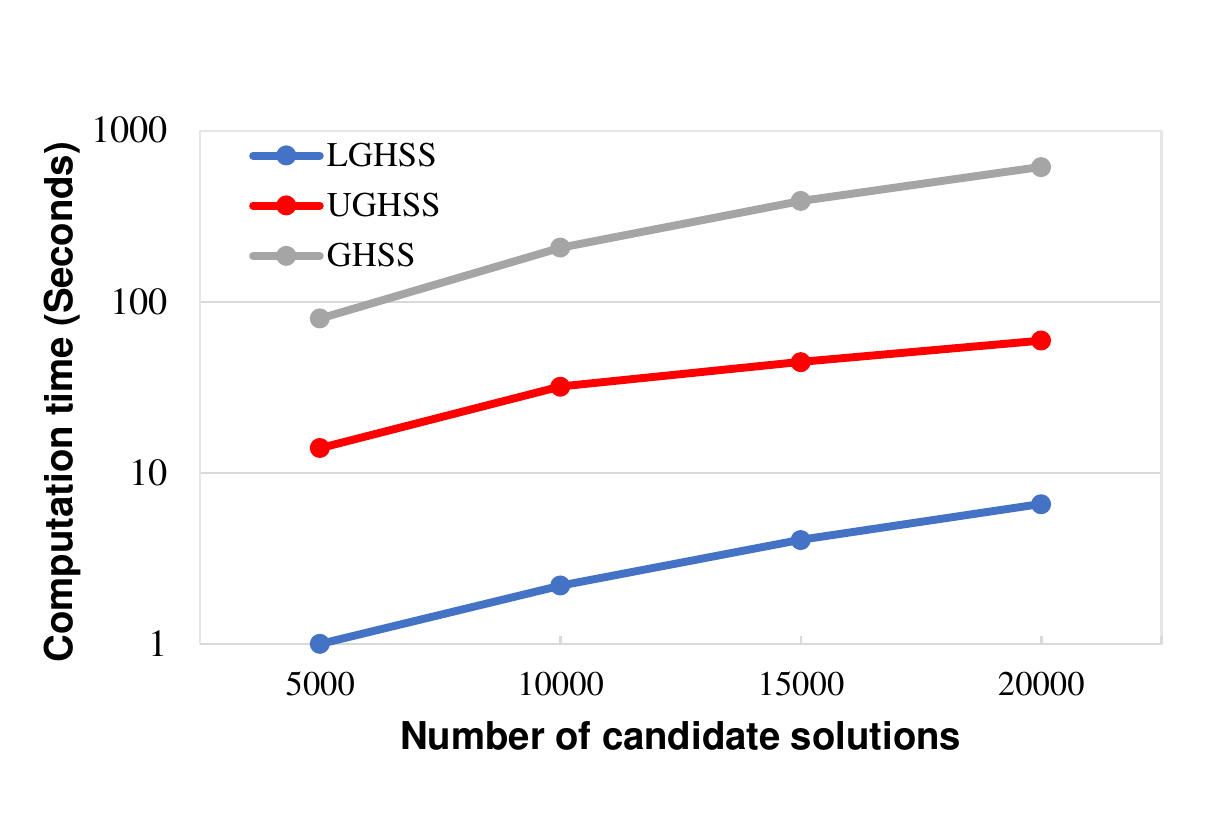}}
 
 \caption{ Average computation time on DTLZ7 with the discontinuous PF. The time axis is log scaled. }
\end{figure}

\section{Concluding Remarks}
In this paper, we proposed an efficient greedy inclusion algorithm (LGI-HSS) to select a small number of solutions from a large candidate solution set for hypervolume maximization. The proposed LGI-HSS algorithm is based on the submodular property of the hypervolume indicator. The core idea of LGI-HSS is to use the submodular property to avoid unnecessary hypervolume contribution calculation. The same solution subset selection result is obtained by LGI-HSS as the standard greedy inclusion algorithm since our algorithm does not change the basic framework of greedy inclusion. Our experimental results on three test problems (DTLZ2, DTLZ7 and Inverted DTLZ2) with 5, 8 and 10 objectives showed that the proposed LGI-HSS algorithm is much more efficient than the standard greedy inclusion algorithm and the state-of-the-art fast greedy inclusion algorithm. 

Our experimental results clearly showed that the idea of lazy evaluation based on the submodular property drastically decreased the computation time of hypervolume-based greedy subset selection. One interesting future research topic is to examine the applicability of this idea to other performance indicators. In this research direction, the relation between the submodularity and the Pareto compliance may need to be clearly explained. Another interesting research direction is to examine the relation between the efficiency of hypervolume-based subset selection algorithms and the properties of multi-objective optimization problems. It needs to be further explained why the increase in the number of objectives did not increase the computation time of some subset selection algorithms for some problems whereas it severely increased for other problems.

\bibliographystyle{IEEEtran}
\bibliography{ref}

\begin{thebibliography}{10}
\providecommand{\url}[1]{#1}
\csname url@samestyle\endcsname
\providecommand{\newblock}{\relax}
\providecommand{\bibinfo}[2]{#2}
\providecommand{\BIBentrySTDinterwordspacing}{\spaceskip=0pt\relax}
\providecommand{\BIBentryALTinterwordstretchfactor}{4}
\providecommand{\BIBentryALTinterwordspacing}{\spaceskip=\fontdimen2\font plus
\BIBentryALTinterwordstretchfactor\fontdimen3\font minus
  \fontdimen4\font\relax}
\providecommand{\BIBforeignlanguage}[2]{{%
\expandafter\ifx\csname l@#1\endcsname\relax
\typeout{** WARNING: IEEEtran.bst: No hyphenation pattern has been}%
\typeout{** loaded for the language `#1'. Using the pattern for}%
\typeout{** the default language instead.}%
\else
\language=\csname l@#1\endcsname
\fi
#2}}
\providecommand{\BIBdecl}{\relax}
\BIBdecl

\bibitem{1li2019empirical}
M.~Li and X.~Yao, ``An empirical investigation of the optimality and
  monotonicity properties of multiobjective archiving methods,'' in
  \emph{International Conference on Evolutionary Multi-Criterion
  Optimization}.\hskip 1em plus 0.5em minus 0.4em\relax Springer, 2019, pp.
  15--26.

\bibitem{2ishibuchi2016compare}
H.~Ishibuchi, Y.~Setoguchi, H.~Masuda, and Y.~Nojima, ``How to compare
  many-objective algorithms under different settings of population and archive
  sizes,'' in \emph{2016 IEEE Congress on Evolutionary Computation
  (CEC)}.\hskip 1em plus 0.5em minus 0.4em\relax IEEE, 2016, pp. 1149--1156.

\bibitem{3tanabe2017benchmarking}
R.~Tanabe, H.~Ishibuchi, and A.~Oyama, ``Benchmarking multi-and many-objective
  evolutionary algorithms under two optimization scenarios,'' \emph{IEEE
  Access}, vol.~5, pp. 19\,597--19\,619, 2017.

\bibitem{4singh2018distance}
H.~K. Singh, K.~S. Bhattacharjee, and T.~Ray, ``Distance-based subset selection
  for benchmarking in evolutionary multi/many-objective optimization,''
  \emph{IEEE Transactions on Evolutionary Computation}, vol.~23, no.~5, pp.
  904--912, 2018.

\bibitem{5bader2011hype}
J.~Bader and E.~Zitzler, ``{HypE: A}n algorithm for fast hypervolume-based
  many-objective optimization,'' \emph{Evolutionary Computation}, vol.~19,
  no.~1, pp. 45--76, 2011.

\bibitem{6bringmann2014generic}
K.~Bringmann, T.~Friedrich, and P.~Klitzke, ``Generic postprocessing via subset
  selection for hypervolume and epsilon-indicator,'' in \emph{International
  Conference on Parallel Problem Solving from Nature}.\hskip 1em plus 0.5em
  minus 0.4em\relax Springer, 2014, pp. 518--527.

\bibitem{7kuhn2016hypervolume}
T.~Kuhn, C.~M. Fonseca, L.~Paquete, S.~Ruzika, M.~M. Duarte, and J.~R.
  Figueira, ``Hypervolume subset selection in two dimensions: Formulations and
  algorithms,'' \emph{Evolutionary Computation}, vol.~24, no.~3, pp. 411--425,
  2016.

\bibitem{8guerreiro2017computing}
A.~P. Guerreiro and C.~M. Fonseca, ``Computing and updating hypervolume
  contributions in up to four dimensions,'' \emph{IEEE Transactions on
  Evolutionary Computation}, vol.~22, no.~3, pp. 449--463, 2017.

\bibitem{9bringmann2014two}
K.~Bringmann, T.~Friedrich, and P.~Klitzke, ``Two-dimensional subset selection
  for hypervolume and epsilon-indicator,'' in \emph{Proceedings of the 2014
  Annual Conference on Genetic and Evolutionary Computation}, 2014, pp.
  589--596.

\bibitem{10rote2016selecting}
G.~Rote, K.~Buchin, K.~Bringmann, S.~Cabello, and M.~Emmerich, ``Selecting k
  points that maximize the convex hull volume,'' in \emph{Proceedings of the
  19th Japan Conference on Discrete and Computational Geometry, Graphs, and
  Games}, 2016, pp. 58--60.

\bibitem{11bradstreet2008fast}
L.~Bradstreet, L.~While, and L.~Barone, ``A fast incremental hypervolume
  algorithm,'' \emph{IEEE Transactions on Evolutionary Computation}, vol.~12,
  no.~6, pp. 714--723, 2008.

\bibitem{12cox2016improving}
W.~Cox and L.~While, ``Improving the {IWFG} algorithm for calculating
  incremental hypervolume,'' in \emph{2016 IEEE Congress on Evolutionary
  Computation (CEC)}.\hskip 1em plus 0.5em minus 0.4em\relax IEEE, 2016, pp.
  3969--3976.

\bibitem{13jiang2014simple}
S.~Jiang, J.~Zhang, Y.-S. Ong, A.~N. Zhang, and P.~S. Tan, ``A simple and fast
  hypervolume indicator-based multiobjective evolutionary algorithm,''
  \emph{IEEE Transactions on Cybernetics}, vol.~45, no.~10, pp. 2202--2213,
  2014.

\bibitem{14nemhauser1978analysis}
G.~L. Nemhauser, L.~A. Wolsey, and M.~L. Fisher, ``An analysis of
  approximations for maximizing submodular set functions-{I},''
  \emph{Mathematical Programming}, vol.~14, no.~1, pp. 265--294, 1978.

\bibitem{15knowles2003bounded}
J.~D. Knowles, D.~W. Corne, and M.~Fleischer, ``Bounded archiving using the
  lebesgue measure,'' in \emph{The 2003 Congress on Evolutionary Computation,
  2003. CEC'03.}, vol.~4.\hskip 1em plus 0.5em minus 0.4em\relax IEEE, 2003,
  pp. 2490--2497.

\bibitem{16zitzler1998multiobjective}
E.~Zitzler and L.~Thiele, ``Multiobjective optimization using evolutionary
  algorithms-a comparative case study,'' in \emph{International Conference on
  Parallel Problem Solving from Nature}.\hskip 1em plus 0.5em minus 0.4em\relax
  Springer, 1998, pp. 292--301.

\bibitem{17bringmann2010approximating}
K.~Bringmann and T.~Friedrich, ``Approximating the volume of unions and
  intersections of high-dimensional geometric objects,'' \emph{Computational
  Geometry}, vol.~43, no. 6-7, pp. 601--610, 2010.

\bibitem{18durillo2010jmetal}
J.~J. Durillo, A.~J. Nebro, and E.~Alba, ``The jmetal framework for
  multi-objective optimization: Design and architecture,'' in \emph{IEEE
  Congress on Evolutionary Computation}.\hskip 1em plus 0.5em minus 0.4em\relax
  IEEE, 2010, pp. 1--8.

\bibitem{19durillo2011jmetal}
J.~J. Durillo and A.~J. Nebro, ``jmetal: A java framework for multi-objective
  optimization,'' \emph{Advances in Engineering Software}, vol.~42, no.~10, pp.
  760--771, 2011.

\bibitem{20}
M.~H. {Overmars} and C.~{Yap}, ``New upper bounds in klee's measure problem,''
  in \emph{[Proceedings 1988] 29th Annual Symposium on Foundations of Computer
  Science}, Oct 1988, pp. 550--556.

\bibitem{21overmars1991new}
M.~H. Overmars and C.-K. Yap, ``New upper bounds in klee's measure problem,''
  \emph{SIAM Journal on Computing}, vol.~20, no.~6, pp. 1034--1045, 1991.

\bibitem{22beume2009s}
N.~Beume, ``S-metric calculation by considering dominated hypervolume as klee's
  measure problem,'' \emph{Evolutionary Computation}, vol.~17, no.~4, pp.
  477--492, 2009.

\bibitem{23while2011fast}
L.~While, L.~Bradstreet, and L.~Barone, ``A fast way of calculating exact
  hypervolumes,'' \emph{IEEE Transactions on Evolutionary Computation},
  vol.~16, no.~1, pp. 86--95, 2011.

\bibitem{24bringmann2009approximating}
K.~Bringmann and T.~Friedrich, ``Approximating the least hypervolume
  contributor: {NP}-hard in general, but fast in practice,'' in
  \emph{International Conference on Evolutionary Multi-Criterion
  Optimization}.\hskip 1em plus 0.5em minus 0.4em\relax Springer, 2009, pp.
  6--20.

\bibitem{25bradstreet2009new}
L.~Bradstreet, L.~While, and L.~Barone, ``A new way of calculating exact
  exclusive hypervolumes,'' \emph{The University of Western Australia, School
  of Computer Science \& Software Engineering, Technical Report
  UWA-CSSE-09--002}, 2009.

\bibitem{31bringmann2010efficient}
K.~Bringmann and T.~Friedrich, ``An efficient algorithm for computing
  hypervolume contributions,'' \emph{Evolutionary Computation}, vol.~18, no.~3,
  pp. 383--402, 2010.

\bibitem{27ulrich2012bounding}
T.~Ulrich and L.~Thiele, ``Bounding the effectiveness of hypervolume-based
  ($\mu$+ $\lambda$)-archiving algorithms,'' in \emph{International Conference
  on Learning and Intelligent Optimization}.\hskip 1em plus 0.5em minus
  0.4em\relax Springer, 2012, pp. 235--249.

\bibitem{26minoux1978accelerated}
M.~Minoux, ``Accelerated greedy algorithms for maximizing submodular set
  functions,'' in \emph{Optimization Techniques}.\hskip 1em plus 0.5em minus
  0.4em\relax Springer, 1978, pp. 234--243.

\bibitem{30leskovec2007cost}
J.~Leskovec, A.~Krause, C.~Guestrin, C.~Faloutsos, J.~VanBriesen, and
  N.~Glance, ``Cost-effective outbreak detection in networks,'' in
  \emph{Proceedings of the 13th ACM SIGKDD International Conference on
  Knowledge Discovery and Data Mining}, 2007, pp. 420--429.

\bibitem{28deb2002scalable}
K.~Deb, L.~Thiele, M.~Laumanns, and E.~Zitzler, ``Scalable multi-objective
  optimization test problems,'' in \emph{Proceedings of the 2002 Congress on
  Evolutionary Computation. CEC'02 (Cat. No. 02TH8600)}, vol.~1.\hskip 1em plus
  0.5em minus 0.4em\relax IEEE, 2002, pp. 825--830.

\bibitem{29deb2013evolutionary}
K.~Deb and H.~Jain, ``An evolutionary many-objective optimization algorithm
  using reference-point-based nondominated sorting approach, part {I}: solving
  problems with box constraints,'' \emph{IEEE Transactions on Evolutionary
  Computation}, vol.~18, no.~4, pp. 577--601, 2013.

\end{thebibliography}
\end{document}